\documentclass[journal]{IEEEtran}

\pdfoutput=1

\usepackage[font=small,skip=5pt]{caption}

\usepackage{pgf,pgfarrows,pgfnodes,pgfautomata,pgfheaps}
\usepackage{graphicx}
\usepackage{amsmath}
\usepackage{multicol}
\usepackage{subfigure}
\usepackage{lipsum}
\usepackage[T1]{fontenc}
\usepackage[ruled]{algorithm2e}
\usepackage{amsfonts} 
\usepackage{gensymb} 
\usepackage{float} 
\usepackage{dblfloatfix}

\title{\fontsize{23.9}{27}\selectfont An Artificial Intelligence System for Combined Fruit Detection and Georeferencing, Using RTK-Based Perspective Projection in Drone Imagery}

\date{}

\begin{document}

\author{Angus Baird and Stefano Giani}

\IEEEaftertitletext{\vspace{-1\baselineskip}}

\markboth{January 2021}{\thepage}
\maketitle

\fontsize{9.5}{11.14}\selectfont

\begin{abstract}

This work presents an Artificial Intelligence (AI) system, based on the Faster Region-Based Convolution Neural Network (Faster R-CNN) framework, which detects and counts apples from oblique, aerial drone imagery of giant commercial orchards. To reduce computational cost, a novel precursory stage to the network is designed to preprocess raw imagery into cropped images of individual trees. Unique geospatial identifiers are allocated to these using the perspective projection model. This employs Real-Time Kinematic (RTK) data, Digital Terrain and Surface Models (DTM and DSM), as well as internal and external camera parameters. The bulk of experiments however focus on tuning hyperparameters in the detection network itself. Apples which are on trees and apples which are on the ground are treated as separate classes. A mean Average Precision (mAP) metric, calibrated by the size of the two classes, is devised to mitigate spurious results. Anchor box design is of key interest due to the scale of the apples. As such, a \textbf{$k$}-means clustering approach, never before seen in literature for Faster R-CNN, resulted in the most significant improvements to calibrated mAP. Other experiments showed that the maximum number of box proposals should be 225; the initial learning rate of 0.001 is best applied to the adaptive RMS Prop optimiser; and ResNet 101 is the ideal base feature extractor when considering mAP and, to a lesser extent, inference time. The amalgamation of the optimal hyperparameters leads to a model with a calibrated mAP of 0.7627.

\end{abstract}

\begin{figure*}[h]
    \centering
    \includegraphics[scale = 0.17]{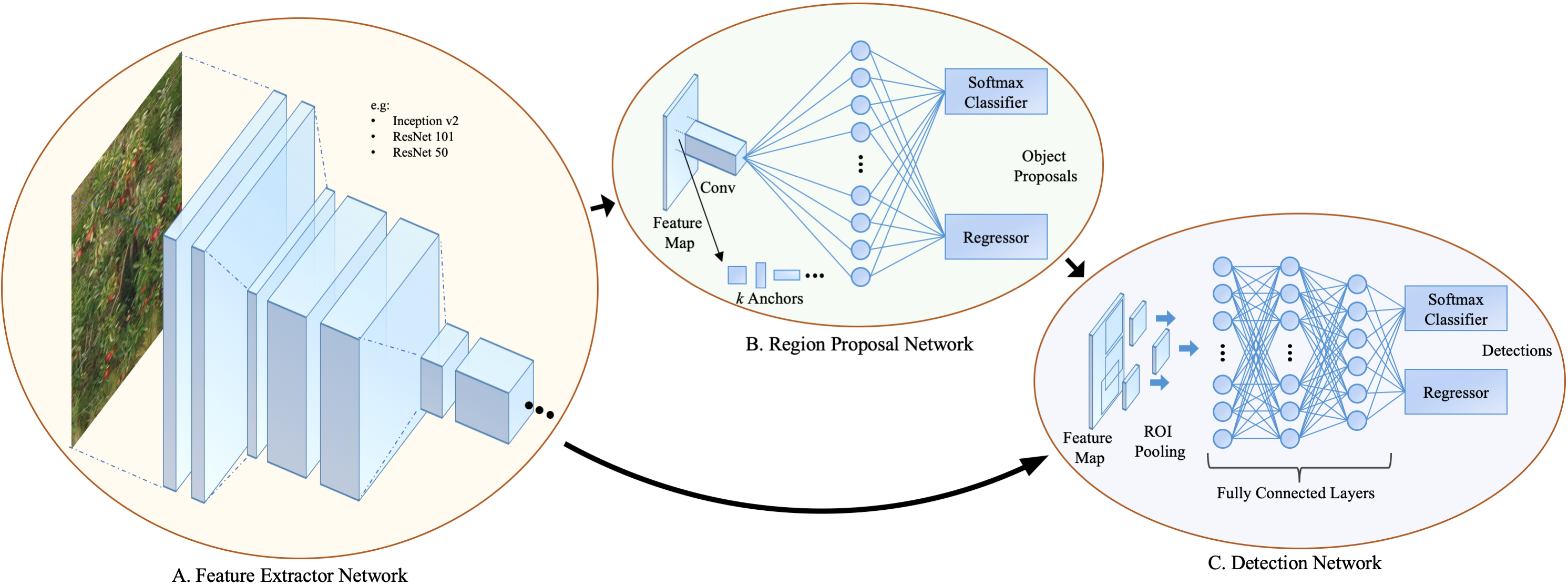}
    \caption{The Faster R-CNN Architecture. Tree-Level Images Are Inputted on the Left-Hand Side, and the Three Main Modules Are Labelled With Letters Corresponding to the Subsections in Section \ref{proposedsolutions}.}
    \label{fig:architecture}
\end{figure*}

\section{Introduction}\label{introduction}

\IEEEPARstart{A}{griculture}, which is the main supplier to the food and drinks industry, contributing \pounds28.2bn to the UK's economy annually, is currently being revolutionised by Artificial Intelligence (AI) \cite{agricultureintheuk}. AI has the potential to increase farming output and efficiency, thereby helping to meet the future demand for resources largely due in part to a rapidly rising population. The United Nations Food and Agriculture Organization (FAO) estimates that the world population will increase by another 2 billion people by 2050, and consequently, food production will be required to increase by between 59\% and 98\% \cite{aiinagriculture}. AI presents solutions to current bottlenecks, providing farmers with a quick go-to-market strategy. Modern examples of agricultural technology (agritech) are detailed in \cite{modernagriculturaltechnology} as autopilot tractors, smart irrigation and crop sensors for efficient fertiliser use, amongst others. One of the most recent developments, however, is the use of drones for surveying orchards. Aerial imagery, spectral reflectances and spatial data can all be collected in one overhead pass of a giant commercial orchard. This enables evidence-based planning to influence policies and decisions taken by farmers \cite{eyeintheskydronerevolution}. Cambridge-based agritech start-up, Outfield Technologies, is one of the leading providers of this service in the UK. As such, they have provided oblique, aerial drone imagery for this research.

The aim of this work is to develop an AI system to detect fruits, in this case apples, from this drone imagery. The desired outputs are the total number of apples in the entire orchard and the number of apples on each individual tree. Apples which are growing on trees and apples which have fallen on the ground will be considered as two separate classes to be counted. 
This distinction contributes to a more accurate and informative yield estimation for the entire orchard. Advanced knowledge of this is beneficial for farmers when making bulk order arrangements with supermarkets. Furthermore, the number of crates and the size of the workforce that will be needed to harvest, box and deliver the fruit can be determined.
To achieve this, the proposed AI system will be based on fundamental principles of multiple-instance object detection. This computer vision technique is the combination of image classification and object localisation, where object coordinates in an image are extracted using bounding boxes with accompanying class labels and confidences. Traditionally, the pipeline of object detection models can be characterised by three stages: informative region selection, feature extraction and classification \cite{detectionreview}.
Conventional methods involved exhaustive sliding window approaches, which were computationally excessive. These were coupled with bespoke feature extractor and classifier pairs, which had to be manually designed for specific types of object \cite{haar, hog}.

Generalisation began following the advent of the Deep Convolutional Neural Network (Deep CNN) which was first used in the 2012 ImageNet Large-Scale Visual Recognition Challenge (ILSVRC). It achieved state-of-the-art accuracy with a top-5 test error rate of 15.3\%, compared to 26.2\% achieved by the second-best entry \cite{imagenet}. This was the pinnacle of a series of iterations on the idea of multi-layer CNNs beginning in 1989 \cite{convnet}.
Modern convolutional object detectors can be separated into two types of framework. The first generates region proposals, then classifies each proposal into different object categories. The second treats object detection as a regression or classification problem \cite{detectionreview}.
Both perform better than the aforementioned traditional approaches due to increased depth, allowing for a greater capacity to learn more complex features and an ability to learn object representations without the need to design features manually \cite{deeplearning}. 
Regions with CNN features (R-CNN) was one of the first examples of CNN-based detection in literature \cite{rcnn}. Adopting the first of the two frameworks, R-CNN applies a selective search to generate 2000 externally computed region proposal crops from an input image \cite{selectivesearch}. Each region is warped and inputted to CNNs \cite{selectivesearch}. Pre-trained, category specific, linear Support Vector Machines (SVM) for multiple classes assign a score to a set of positive and negative regions. These are then adjusted using bounding box regression and filtered with Non-Maximum Suppression (NMS), resulting in final bounding boxes for preserved object locations \cite{detectionreview}.
The excessive computational cost associated with extracting and subsequently passing each region through a CNN was solved in Fast R-CNN. This passes the input image through a single CNN, creating a convolutional feature map. From this, region proposals are obtained and reshaped using a Region of Interest (ROI) pooling layer before being fed into a fully connected network \cite{fastrcnn}.
A bottleneck in this method becomes the computation of region proposals. Faster R-CNN evolved from this, showing that box proposals can be generated using a Neural Network (NN). This additional Region Proposal Network (RPN) was introduced using a fully convolutional NN that shares full-image convolutional features with the detection network, thus enabling nearly cost-free region proposals \cite{fasterrcnn}. The network slides over the convolutional feature map and, at each window, generates an array of anchor boxes of different shapes and sizes. Rectified Linear Unit (ReLU) activation is applied to the output of the convolutional layer to increase non-linearity. The regression towards true bounding boxes is achieved by comparing proposals relative to anchors.
Aside from the R-CNN family, other architectures using the first kind of framework include Region-Based Fully Convolutional Network (R-FCN) which is fully convolutional with almost all computation shared on the entire image as opposed to applying a per-region sub-network hundreds of times. To achieve this, position-sensitive score maps address a dilemma between translation-invariance in image classification and translation-variance in object detection \cite{rfcn}. 
The most ubiquitous networks using the second kind of detection framework are You Only Look Once (YOLO) \cite{yolo}, its successors, YOLOv2 and YOLOv3 \cite{yolov2,yolov3NEW} and Single Shot Multibox Detection (SSD) \cite{ssd}. All of which are optimised for real-time applications, due to the lack of region proposal generation.

An accurate detector is prioritised for this application due to the inherent difficulties of the detection task. These include: the small size of the apples, some being in the range a few pixels in area, but with varying scales and aspect ratios; the fact that many of the apples are largely concealed by other overlapping apples, or are partially covered by foliage; the poor capturing of some apples due to motion blur in the done camera; and that the number of instances of apple in each photo is very high.

This paper first justifies the choice of object detection framework. Then, a range of different feature extractor networks are proposed and subsequently investigated. These form the basis upon which relevant hyperparameters are tuned for their effect on the accuracy
and, to a lesser extent, inference time. This culminates in the proposed optimal model configuration, which features a combination of all the most successful hyperparameter modifications. 
Lastly, a novel precursory system is presented as part of the practical deployment of the network. This serves as an image preprocessor, which crops raw drone imagery into uniquely labelled crops of each tree. This enables yield information to be localised to the tree-level. In doing so, this prevents double counting of apples and results in yield representation at the tree-level, as opposed to merely a total for the number of apples in the whole orchard.
The final output consists of this prepossessing stage and the trained optimal model configuration. Supplementary material, including code for dataset augmentation, amongst others, are referred to in the report.

\section{Proposed Solutions}\label{proposedsolutions}

To justify the type of object detection framework used, the regression and classification approaches, such as YOLOv2, YOLOv3 and SSD are first considered \cite{yolov2,yolov3NEW,ssd}. These have been shown to achieve similar mean Average Precision (mAP) as the region proposal-based networks on the benchmark PASCAL VOC 2012 dataset \cite{mobilenet}. This can be misleading of overall performance, especially relating to this application, due to the lack of small objects in the dataset. A more thorough test is the Microsoft Common Objects in Context (COCO) dataset. Average Precision (AP) performance of SSD on small ($<32^2$ pixels) and medium ($32^2<96^2$ pixels) objects (using COCO definitions) is on average $66.7\%$ and $33.9\%$ less than the corresponding Faster R-CNN, respectively \cite{comparisonpaper}. This was caused by large localisation errors due to an inability to fully represent the characteristics of smaller objects in particular \cite{smallobjectdetection}. Similar is found with YOLO architectures.

Considering the region proposal approaches, R-FCN, despite its region-based detector being fully convolutional, has been shown to be able to match Faster R-CNN on all three object size categories. When using feature extractors with more layers however, comparative accuracy was found to tail off in all size categories by approximately equal proportions \cite{comparisonpaper}.
Therefore, due to accuracy taking priority over inference time, and the amount of small and medium apple objects in this dataset, there is sufficient justification for Faster R-CNN to be selected as the object detection framework. 

Figure \ref{fig:architecture} illustrates the proposed Faster R-CNN architecture. This is partitioned into three main modules for the purposes of this work: the initial CNN layers for feature extraction; the RPN, which outputs a set of object proposals with a corresponding objectness score; and the final ROI pooling and fully connected layers, which are comparable to Fast R-CNN \cite{fastrcnn}.

\subsection{Feature Extractor Network}

The CNN architecture for feature extraction can be effectively decoupled from the rest of the network, which will be referred to hereafter as the Faster R-CNN meta-architecture. While most application-specific literature simply uses the base networks presented in the original paper, a more modular approach to tuning hyperparameters is taken here to achieve the optimal solution. This is due to the intertwined nature of Faster R-CNN, and the meta-architecture depending upon a rich feature map. Moreover, this approach has been shown to yield positive results in the pertinent work \cite{remotesensingimagery} which looks at multi-scale object detection in remote sensing imagery.

The feature extractors that will form the basis of the investigation are Inception v2 and ResNet (including the 50-layer and 101-layer variants). These have been shown to achieve top accuracy (with a mAP of 0.28, 0.30 and 0.32, respectively) when trained and tested on the COCO dataset, within reasonable time, using a Nvidia GeForce GTX TITAN X card \cite{modelzoo}. The mAP data however are only reported as an average over Intersection over Union (IoU) thresholds in the range 0.50 to 0.95 at increments of 0.05. Such a metric is excessively strict for the application of this paper, but serves as a good guide nonetheless. Moreover, performance of these feature extractors on the small object category also remains unknown.
Later in the work, the combined Inception ResNet v2 Atrous network is examined due to a reported mAP of 0.37, at the expense of being between 6 and 11 times slower than the separated feature extractor networks at inference time.

\subsubsection{Inception}
\begin{figure}[H]
    \centering
    \includegraphics[scale = 0.26]{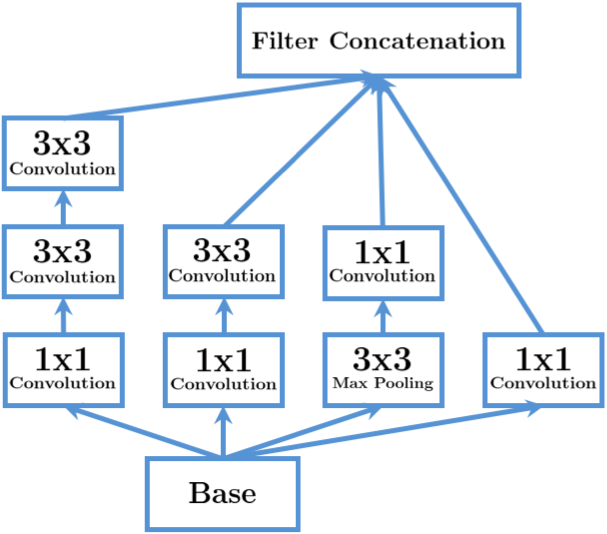}
    \caption{The Inception Block: Split, Transform and Merge Concept.}
    \label{fig:inception_module}
\end{figure}

Due to variation in location and scale of apples in the images, optimising filter size is difficult. Moreover, excessively deep networks are more susceptible to overfitting and vanishing gradient. A solution is to increase the width of the network, instead of the depth, and have multiple-sized filters operating on the same level. The Inception block incorporates multi-scale convolutional transformations on the premise of split, transformation and concatenation \cite{reviewpaperpakistan}. This permits branching within a layer, allowing for abstraction of features at different spatial scales \cite{inception}. Inception v2 was later proposed to reduce representational bottleneck caused by the reduction in dimensions due to convolutions \cite{inceptionv2}. An Inception block, which is found in the Inception v2 network used in this work, is shown in Figure \ref{fig:inception_module}.

\subsubsection{ResNet} 
The aforementioned problem of vanishing gradient is also addressed by skip connections. These jump over layers, reusing activations from the previous layer until the adjacent layer learns its weights \cite{resnet}. The mathematical formulation for this is given by:

\begin{equation}
\pmb{F}_{m+1}^{k^\prime} = g_{c} ( \pmb{F}_{l \rightarrow m}^k , \pmb{k}_{l \rightarrow m} ) + \pmb{F}_{l}^k, \; \; \; \; \; \; m\ge l
\label{eq:resnet1}
\end{equation}

\begin{equation}
\pmb{F}_{m+1}^k = g_{a} ( \pmb{F}_{m+1}^{k^\prime} ),
\label{eq:resnet2}
\end{equation}

\begin{equation}
g_{c} ( \pmb{F}_{l \rightarrow m}^k , \pmb{k}_{l \rightarrow m} ) = \pmb{F}_{m+1}^{k^\prime} - \pmb{F}_{l}^k,
\label{eq:resnet3}
\end{equation}

where $g_{c} ( \pmb{F}_{l \rightarrow m}^k , \pmb{k}_{l \rightarrow m} )$ is a transformed signal, and $\pmb{F}_{l}^k$ is an input of the $l^{th}$ layer. In Equation \ref{eq:resnet1}, $\pmb{k}_{l \rightarrow m}$ shows the $k^{th}$ processing unit, whereas $l \rightarrow m$ suggests that the Residual block can be consistent of one or more hidden layers. Original input $\pmb{F}_{l}^k$ is added to the transformed signal through the bypass pathway and thus results in an aggregated output, $\pmb{F}_{m+1}^{k^\prime}$, which is assigned to the next layer after applying activation function, $g_{a}$. Whereas, $\pmb{F}_{m+1}^{k^\prime} - \pmb{F}_{l}^k$ returns residual information which is used to perform reference-based optimisation of weights. A further discussion of this architecture is summarised in \cite{reviewpaperpakistan}, which maintains the same notation as used in this paper.

\begin{figure}[H]
    \centering
    \includegraphics[scale = 0.21]{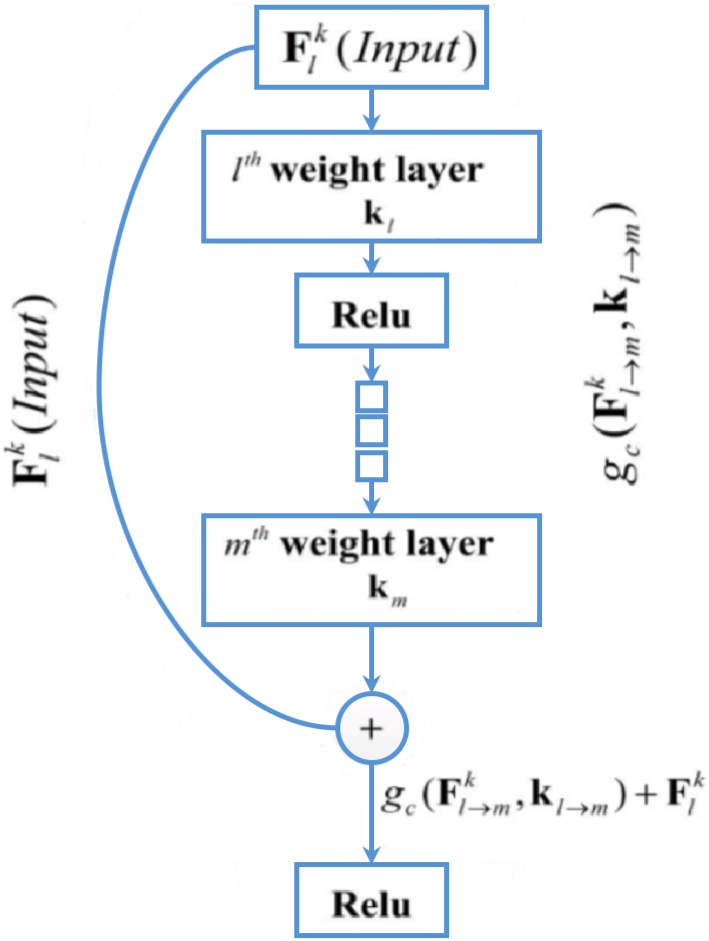}
    \caption{The Residual Block: ResNet Structural Unit \cite{reviewpaperpakistan}.}
    \label{fig:resnet_module}
\end{figure}

This type of residual learning is shown by the Residual block in Figure \ref{fig:resnet_module}. This is a structural unit of the ResNet family of architectures. This work looks at two of the most common variants: ResNet 50 and ResNet 101, which are the 50-layer and 101-layer architectures. The effect of network depth on eventual detection performance can therefore be ascertained.

\subsubsection{Inception ResNet}
This architecture combines the power of residual learning and the Inception block. In doing so, the filter concatenation stage is replaced by the Residual block \cite{reviewpaperpakistan} \cite{inceptionresnet}. The specific incarnation looked at in this paper, Inception ResNet v2 Atrous, uses atrous convolutions to maintain field of view at each network layer. This was reported to outperform other varieties of the network \cite{modelzoo}.

\subsection{Region Proposal Network}
The salient feature of Faster R-CNN is the RPN. This is a deep, fully convolutional network that proposes the regions in a nearly cost-free way, by sharing full image convolutional features, as well as weights, with the rest of the network \cite{detectionreview}. 

\subsubsection{Anchor Boxes}
The RPN slides over the feature map, outputted at the last shared convolutional layer of the feature extractor network. For each pixel, a set of $k$ anchors, spanning the input image, are overlaid, each with a set of aspect ratios and scales. The network checks if an object is contained within these, or it refines the coordinates of the anchors to give bounding boxes as object proposals \cite{objectdetectionatechnicalsummary}. 

\subsubsection{Loss Function}
Training therefore consists of making two predictions for each anchor: a discrete class prediction and a continuous prediction of an offset by which the anchor needs to be translated in order to encompass the groundtruth bounding box \cite{comparisonpaper}. There is therefore a multi-task loss consisting of classification loss and bounding box regression loss. When training the RPN, a binary class label is assigned to each anchor. As in most literature, this work maintains the convention that positive labels are given to anchors with an IoU overlap greater than 0.7 with the groundtruth box. Conversely, negative labels are assigned to non-positive anchors if the IoU overlap is lower than 0.3 for all groundtruth boxes, as in \cite{fasterrcnn}. The loss function to be minimised is given by,

\begin{align}
\begin{split} 
L( \{p_{i}\} , \{t_{i}\} ) =& \frac{1}{N_{cls}} \sum_{i} L_{cls} ( p_{i} , p_{i}^{\star} )\\
+& \lambda \frac{1}{N_{reg}} \sum_{i} p_{i}^{\star} L_{reg} ( t_{i}, t_{i}^\star ).
\label{eq:loss_function}
\end{split}
\end{align}

In this, $p_{i}$ is the predicted probability of the $i$-th anchor being an object. The groundtruth label, $p_{i}^{\star}$, is 1 for positive anchors and 0 otherwise. The vector, $t_{i}$, represents 4 parameterised coordinates of the predicted bounding box; $t_{i}^{\star}$ is related to the groundtruth box associated with a positive anchor. $L_{cls}$ is a binary log loss and $L_{reg}$ is a smoothed $L_{1}$ loss as defined in \cite{fastrcnn}. These two terms are normalised by the mini-batch size, $N_{cls}$, and the number of anchor locations, $N_{reg}$, respectively. $\lambda$ is a balancing parameter which is set to 10 in this work, as was found to maximise mAP in the original paper \cite{fasterrcnn}.

\subsection{Detection Network}

The final part of the meta-architecture is a network, equivalent to Fast R-CNN, which uses the previously generated proposed regions \cite{fastrcnn}. This network consists of an ROI pooling layer which takes these region proposals from the feature map, divides them into subwindows, then performs max pooling over these subwindows. The output is passed through fully connected layers and the features are fed into the sibling classification and bounding box regression branches. The classification branch uses softmax to convert class scores. The regression branch, which is size agnostic but specific to each class, unlike in the RPN, has coefficients which are used to improve the predicted bounding boxes \cite{objectdetectionatechnicalsummary}.

\section{Experimental Procedure}\label{experimental_procedure}

\subsection{Dataset}

The raw dataset consists of oblique drone imagery. Each image is 5472 by 3648 pixels in area, and contains roughly 100 trees. In some images, up to 70\% of the image is background and does not contain trees however. Therefore, to avoid unnecessary computation, it is decided that the input to the feature extractor NN will be crops of individual trees from the raw drone images. Obtaining these crops is an automated process which is further motivated by the desire to pair trees with a unique identifier. This enables a geospatial representation of the yield of fruit at the tree-level. The deployment of this system, as a precursor to the object detection NN, is discussed further in Section \ref{sec:deployment}. 

To create training and validation datasets, the tree-level crops are marked-up manually using \textit{labelimg.py} \cite{labelimg}.
Bounding boxes are drawn for two classes: \textit{tree\_apple} (for apples which are on the trees) and \textit{ground\_apple} (for apples which have fallen to the ground). Distinctions between these two classes include slight discolourations on the surface of ground apples, as well as detectable differences between a tree backdrop and a ground backdrop around the edge of the apples. It must be noted that there exists a degree of subjectivity in the manual mark-up process. In some cases, apples, particularly tree apples, are mostly concealed, blurred, or are clustered in such a way that it is impossible to determine precisely how many bounding boxes should be drawn. Consequently, it is expected, given this and the similarity between the objects in the two classes, that the mAPs found in this paper will be lower than those seen in the literature for more objective object detection applications.

\begin{figure}[H]
    \centering
    \includegraphics[scale = 0.138]{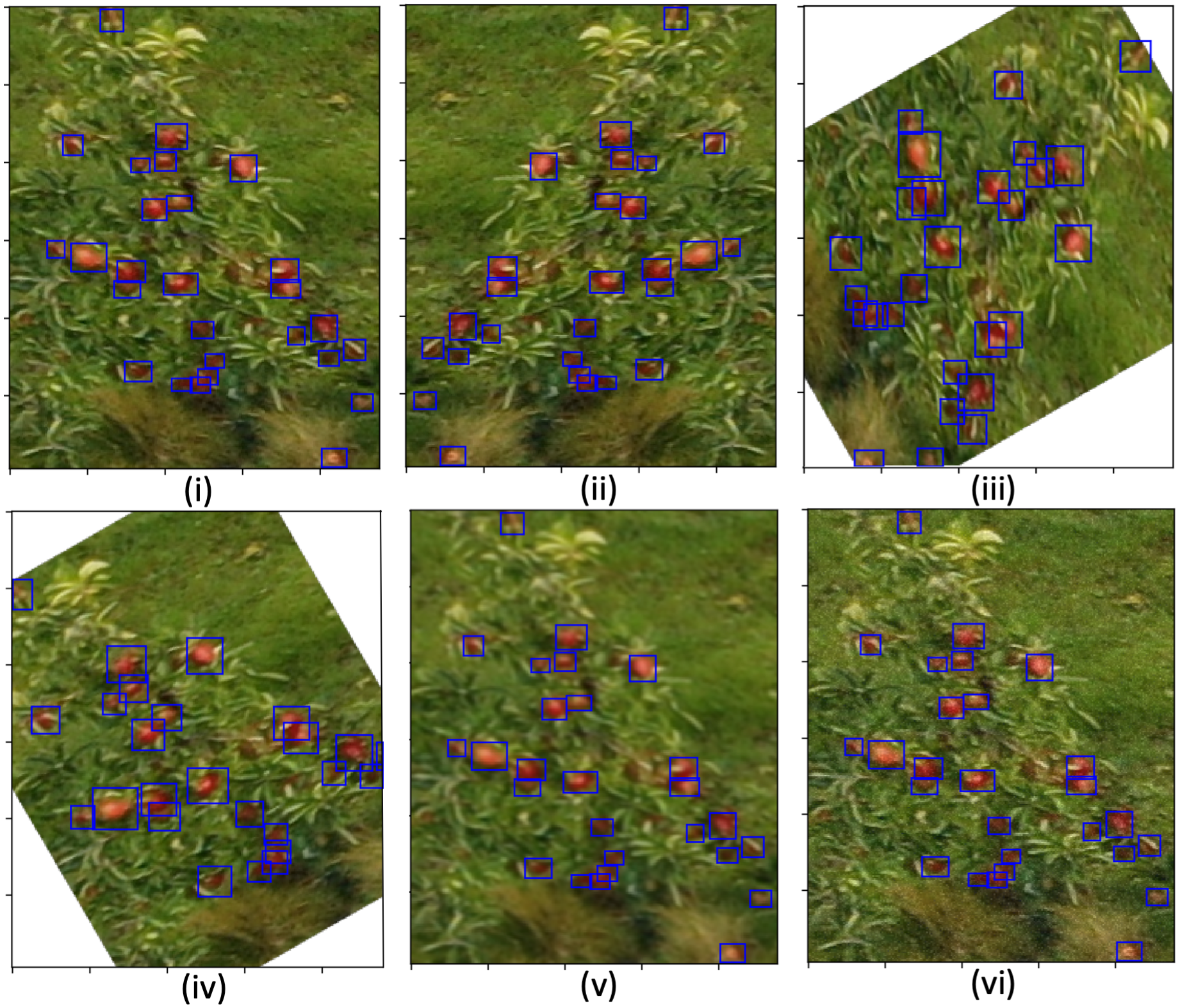}
    \caption{Examples of Data Augmentation: (i) Original, (ii) Mirrored Horizontally, (iii) 60\degree Clockwise Rotation, (iv) 30\degree Anticlockwise Rotation, (v) Gaussian Blur $\sigma$=0.1, (vi) Additive Gaussian Noise From Distribution: N(0, 0.05*255). }
    \label{fig:img_aug}
\end{figure}

To expand the dataset efficiently, data augmentation techniques are applied using \textit{image\_augmentation.py}, which was written largely using the imgaug library. Relevant transformations that preserve the semantic content of the images are explored, including: rotation (up to $\pm60\degree$) and horizontal mirroring, as well as subtle Gaussian blur and additive noise which imitate the occasional motion blur during drone image capture. Examples of these are shown in Figure \ref{fig:img_aug}; the transformed groundtruth bounding boxes are shown in blue.

The traditional 80:20 partition is applied to form the training and testing datasets, respectively. A further 80:20 split on the training dataset separates the images to be used in training, from those used in the validation of the models \cite{pareto}. In total, 13,439 apples are marked-up in 435 tree-level images. 
For the initial purposes of training a large variety of different models however, a subset of approximately a quarter of the data (107 images and 2,709 marked-up apples) is used, due to time constraints. Training with this smaller set is used for preliminary cross-evaluation of different feature extractors and hyperparameter combinations. Later, the effect on accuracy of training with the full dataset is explored. This is also used to train the optimal model configuration.

\subsection{Training, Validation and Testing}

Training, validation and testing of all 76 models used in this work is performed with a Nvidia TITAN Xp card running TensorFlow GPU. The identical 107-image abbreviated dataset is used on all models initially, with final experiments using the full 435-image dataset. All models are trained to 200,000 steps, which has been shown experimentally to be more than sufficient to train the Oxford-IIIT Pets dataset \cite{modelzoo}.
After training, as with most object detectors in literature, validation mAP is measured and compared. This work uses the ubiquitous PASCAL VOC metric, which includes a class-specific AP at IoU = 0.50 as well as a corresponding combined-class mAP. In this work however, a calibrated mAP at IoU = 0.50 is devised as the primary metric. This is weighted by the ratio 92:8 between the two classes, corresponding to the number of instances of the \textit{tree\_apple} class to \textit{ground\_apple} class, respectively. This aims to mitigate sporadic AP readings due to the relatively small number of ground apples, which can over-influence the overall mAP. Lastly, the speed of the models is measured by recording the inference time on a sample of the testing set consisting of 15 images.

The first collection of experiments concerns the baseline models \cite{modelzoo}. These are three models which differ in backbone feature extractor network: Inception v2, ResNet 101 and ResNet 50. The meta-architecture of each baseline model begins the same in terms of hyperparameters. The key values are shown in Table \ref{tab:baseline_params}. Hyperparameters which are not listed can be assumed to be as they are in \cite{fasterrcnn}.

\begin{table}[H]
\begin{center}
\caption{Baseline Hyperparameters}\label{tab:baseline_params}
\begin{tabular}{lll}
\hline \hline
Field & Parameter & Value(s)\\
\hline \hline
Image Resizer & Min Dimension & 600 \\
\tiny{(Fixed Aspect Ratio)} & Max Dimension & 1024 \\
\hline
First Stage Features & Stride Length & 16 \\
\hline
Grid Anchors & Scales & [0.25 0.5 1.0 2.0] \\
 & Aspect Ratios & [0.5 1.0 2.0] \\
 & Height Stride, Width Stride & [16 16] \\
\hline
Object Proposals & NMS IoU Threshold & 0.7 \\
 & Max Box Proposals & 150 \\
\hline
Training Config & Optimiser Type (Value $\gamma$) & Momentum (0.9) \\
 & Initial Learning Rate $\eta$ & 0.0003 \\
\hline \hline
\end{tabular}
\end{center}
\end{table}

For each baseline model, the next set of experiments involve configuring the anchor boxes in the RPN stage, specifically, stride lengths, scales and aspect ratios. Clustering approaches to anchor box design, which have never been seen in literature for Faster R-CNN, are compared to the baseline and original anchor boxes in \cite{fasterrcnn}. Also in the RPN, the number of box proposals will be varied, noting the speed/accuracy compromise. Experiments then focus on the training configuration, in particular, the type of optimiser and the learning rate. Three of the most efficient types are looked at, including the baseline case, Momentum, which is based on Stochastic Gradient Descent (SGD) but with a temporal element, $t$ \cite{optimiserreviewpaper}. This essentially provides momentum for updates on parameter, $\theta$, allowing for faster convergence and reduced oscillations. This is given by,

\begin{equation}
\theta_{t} = \theta_{t} - \eta\nabla L(\theta) + \gamma \nu_{t}
\label{eq:momentum}
\end{equation}

where $\nu_{t}$ is the previous update to $\theta$. Momentum is compared to the Adam optimiser which computes adaptive learning rates for each parameter \cite{adam}. Also considered is another adaptive learning rate optimiser, RMS Prop. This unpublished method divides the learning rate by an exponentially decaying average of squared gradients \cite{optimiserreviewpaper}. After the training configurations are explored, experiments look at combining Inception and ResNet architectures in the backbone feature extractor CNN, followed by quantifying the effect of training on more data. Once these experiments have been run singly, the optimal solution is finally configured by incorporating the best performing hyperparameter combinations from all the previous models.

\section{Results and Discussion}\label{sec_results}

For experimental validity, the identical Inception v2 baseline model configuration was repeatedly trained to establish that the resulting variation in model mAP was sufficiently small to justify further model comparison. From 5 identical models, the mean calibrated mAP was 0.5232 with a variance of $0.005^2$. Certain hyperparameters from Table \ref{tab:baseline_params} which were investigated but did not show any improvement on the baseline models include the input image resizer dimensions and the NMS IoU threshold. Consequently, such experiments are not reported in detail.

\subsection{Baseline Models}

The three baseline models, distinguishable only by their feature extractor networks, were trained first on the 107-image dataset. The results, including class-specific and calibrated validation mAP (at the IoU=0.50 threshold) as well as the test inference time, are given in Table \ref{tab:baseline_results}.   

\begin{table}[H]
\begin{center}
\caption{Baseline Performance Statistics}\label{tab:baseline_results}
\begin{tabular}{lllll}
\hline \hline
Feature & \textit{ground\_apple} & \textit{tree\_apple} & Calibrated & Inference\\
Extractor & AP & AP & mAP & Time (s)\\
\hline \hline
Inception v2 & 0.4091 & 0.5235 & 0.5144 & 16.43 \\
ResNet 101 & 0.2213 & 0.6700 & 0.6341 & 56.38 \\
ResNet 50 & 0.4124 & 0.6160 & 0.5997 & 35.27 \\
\hline \hline
\end{tabular}
\end{center}
\end{table}

The AP for the \textit{tree\_apple} class reflects what was reported in \cite{modelzoo} for these CNNs on COCO. The \textit{ground\_apple} class, on the other hand, does not correlate. This is likely due to the presumed lack of training data for this class in comparison to the \textit{tree\_apple} class. The calibrated mAP is, as expected, found to be more representative of actual accuracy on the test set. It can also be seen that test inference time is proportional to mAP, with the ResNet architectures having the highest calibrated mAP but the longest inference time. Inception v2 was found to be up to 3.5 times faster but with a considerably lower calibrated mAP. Furthermore, doubling the number of ResNet layers yields an increase in calibrated mAP by 5.58\%.

The findings demonstrate that the deeper, residual learning-based architecture of the ResNets is more effective at representing and extracting the small, low-level features of the apples compared to the width-based approach of Inception. Interestingly however, Inception v2 was found to outperform both ResNet architectures, achieving nearly double the mAP, for medium sized ($32^2<96^2$ pixels) apples. Though, due to the infrequency of these apples, this evidently was not reflected in the overall mAP.

\subsection{Anchor Boxes}

The family of tuneable hyperparameters found to be key in improving detection performance was anchor box design. Previous approaches gauge these values using ad-hoc heuristic methods for the chosen application, for instance \cite{textdetectionexample}. Two approaches, never before seen in literature surrounding Faster R-CNN or any other region-based detector networks, are trialed. Both are based on the $k$-means clustering algorithm \cite{kmeansoriginalpaper}: one using Euclidean distance and the other using IoU \cite{yolov2} as the distance metric, $d$, as in

\begin{equation}
d\text{(b box, centroid)} = 1 - \text{IoU(b box, centroid)}.
\label{eq:IoU_metric}
\end{equation}

The two use Exploratory Data Analysis (EDA) of the groundtruth bounding boxes from the training dataset to determine anchor scales and aspect ratios as the $k$-means centroid, $\mu$, of boxes in cluster $\omega$. This is shown in Algorithm 1.













\begin{figure}[H]
    \centering
    \includegraphics[scale = 0.457]{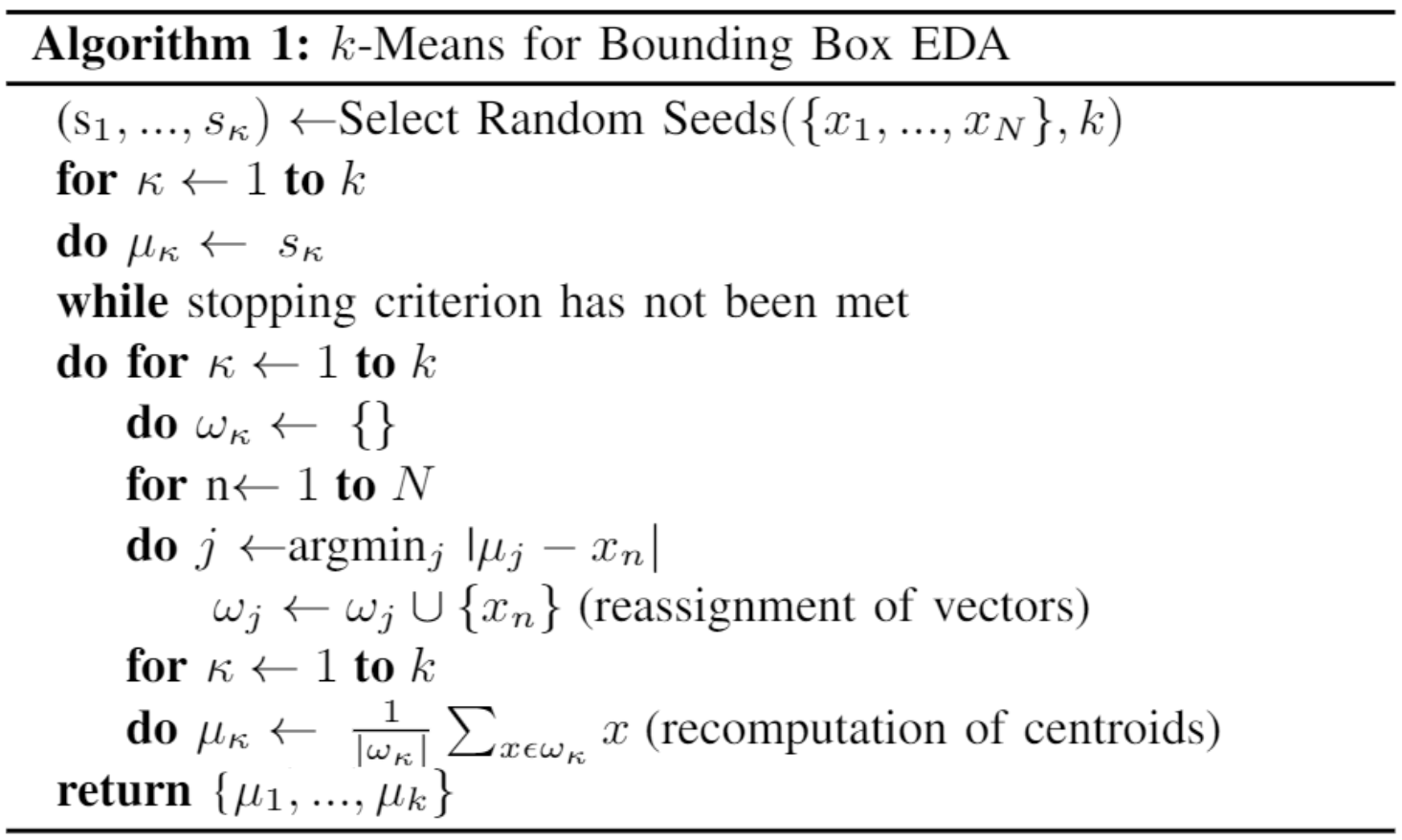}
    \label{fig:algorithm_indent}
\end{figure}

\begin{figure}[H]
    \centering
    \includegraphics[scale = 0.47]{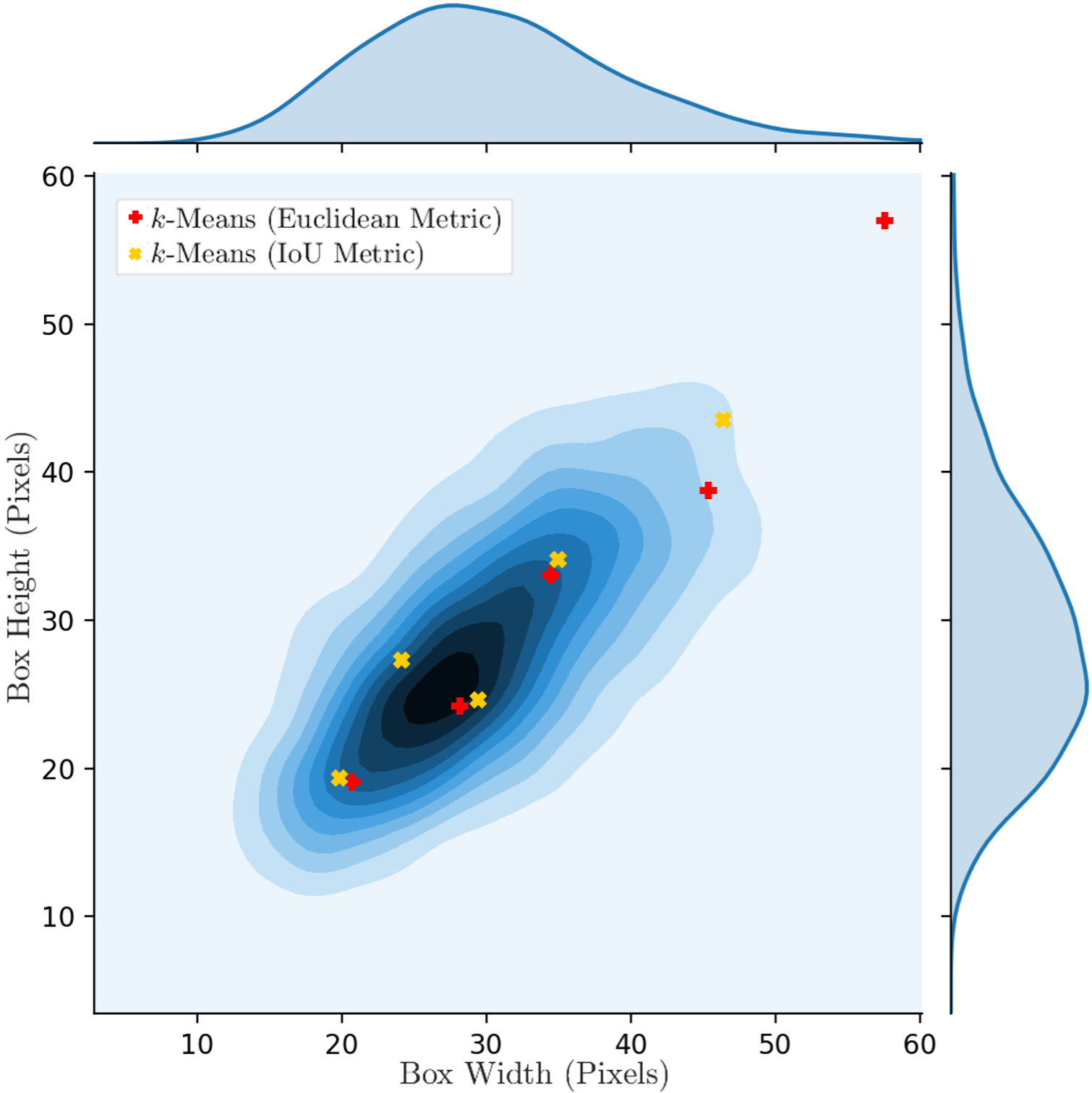}
    \caption{KDE Plot of Groundtruth Bounding Boxes With Overlaid Anchor Designs From the Two $k$-Means Clustering Approaches.}
    \label{fig:EDA_box}
\end{figure}

The Within-Cluster Sum of Squared Errors (WSS) was calculated for $k$ between 1 and 10. Using the elbow method, $k=5$ was found most suitable. The EDA of bounding boxes is shown in Figure \ref{fig:EDA_box} as a Kernel Density Estimate (KDE) plot. The resulting anchor box coordinates of the two $k$-means metrics are superimposed as centroids of the 5 clusters. 

To assess the performance of the $k$-means approaches, a manually selected batch of anchors was chosen with regular interval scales and aspect ratios over the dataset, as was found to be most successful in previous works \cite{comparisonpaper}. Also tested were the 9 anchors which were proposed in the original Faster R-CNN paper for the PASCAL VOC set \cite{fasterrcnn}. For completeness, the baseline anchors are also featured. Table III 
shows all anchor designs.


\begin{figure}[H]
    \centering
    \includegraphics[scale = 0.1005]{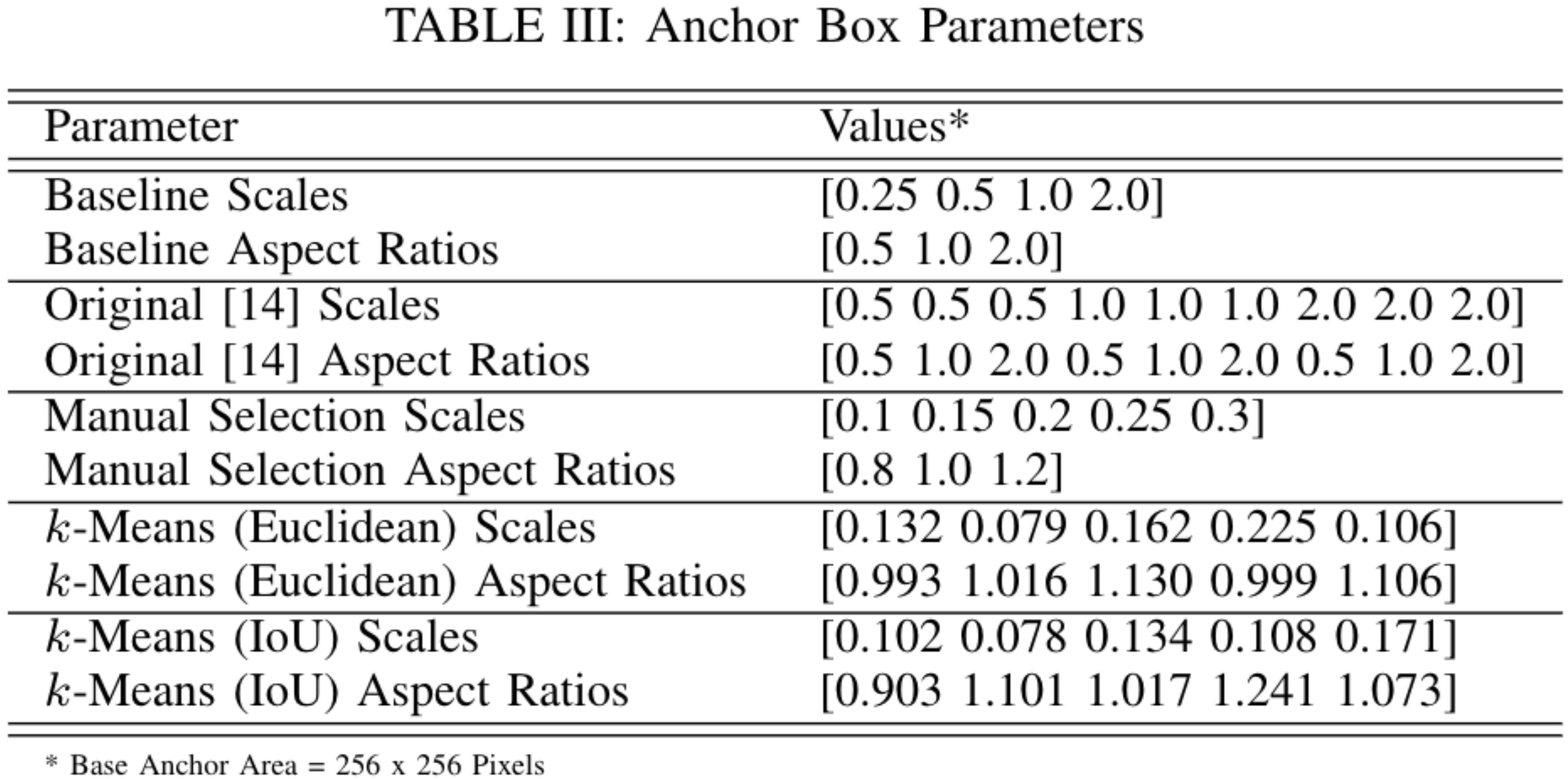}
    \label{fig:table_shrunk}
\end{figure}

\setcounter{table}{3}

\begin{figure}[H]
    \centering
    \includegraphics[scale = 0.272]{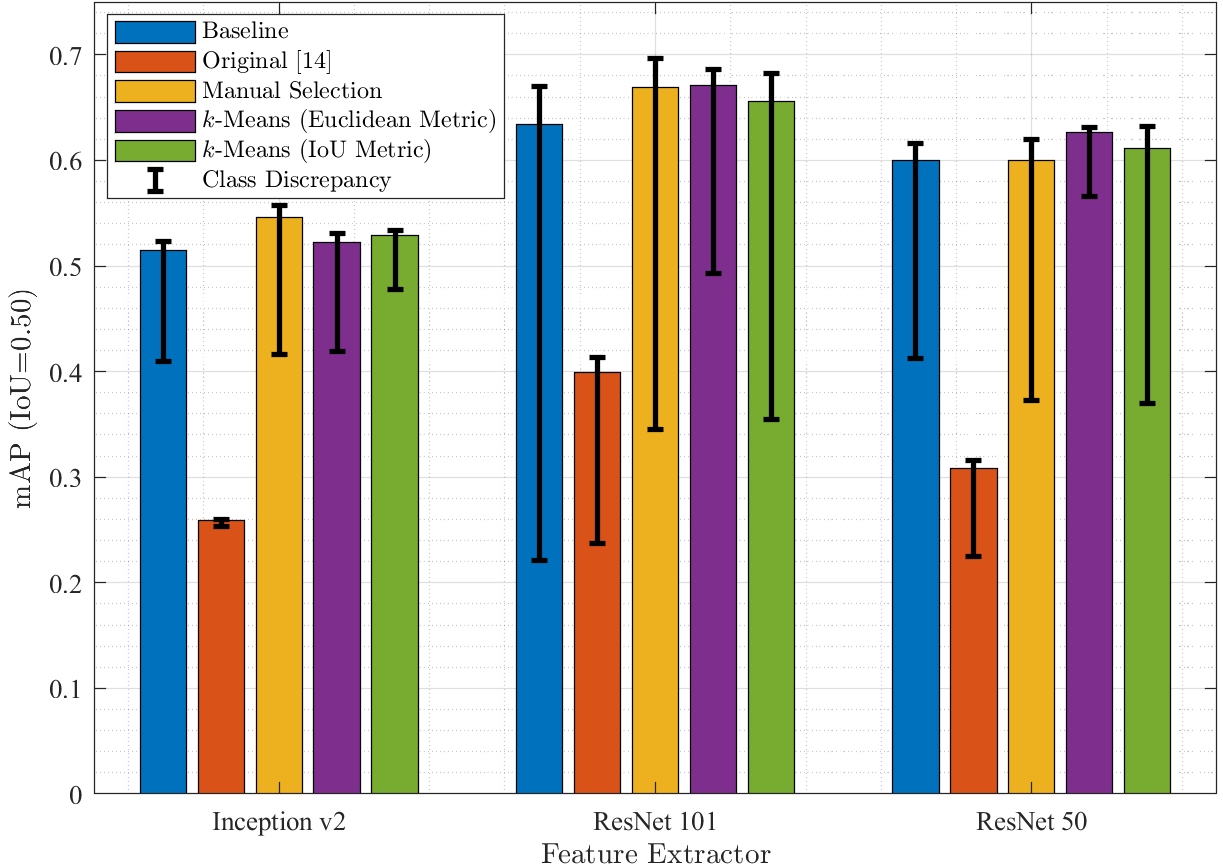}
    \caption{mAPs For Different Anchor Box Design Methodologies. Bars Represent Calibrated mAP. Error Bars Denote the Class Discrepancy: AP of the \textit{tree\_apple} Class (Upper Bound), AP of the \textit{ground\_apple} Class (Lower Bound).}
    \label{fig:anchor_box_bars}
\end{figure}

Experiments on the height and width stride for the anchors, with different combinations of first stage features stride are not reported in detail. When sweeping the first stage features stride down to 8, an insignificant increase in \textit{tree\_apple} AP was recorded, when considering the previously mentioned variance in successive training. The \textit{ground\_apple} AP, on the other hand, was detrimentally reduced. Furthermore, it was found that sweeping anchor height and width strides between 8 and 64 reduced or did not improve the calibrated mAP. This might be explained by the small size and sparsity of apples in most tree-level images, particularly those captured towards the back of the field of view of the drone's camera. Also unreported are the inference times, which were found not to differ significantly from the baselines cases. The mAPs for all cases are shown in Figure \ref{fig:anchor_box_bars}.

Immediately evident is the poor performance of the anchors from the original paper. This shows the importance of optimising anchor shapes for a particular dataset. It also shows that merely having more anchors at each location on the convolutional feature map is insufficient. This is exemplified by the baseline set which, despite using fewer anchors, includes the smaller, 0.25 scale anchor which is more appropriately fitted to the typical instance of apple class. This also contributes to the reduction of false negatives, which is often caused by a bunch or several overlapping apples on a tree being miss-detected as a single instance. This results in an increase in Average Recall (AR). This was also noted in the three approaches which considered the groundtruth bounding box dimensions. These were the best performers on each of the three feature extractors. For Inception v2, the manually chosen anchors yielded a higher calibrated mAP than both $k$-means, but had a larger class discrepancy. This could potentially be explained by a mismatch between the receptive field of the network and the relatively smaller anchor boxes generated by the two $k$-means approaches \cite{receptivefield}. For the two, deeper, ResNet architectures however, the best performing anchor design (highest calibrated mAP and smallest class discrepancy) was obtained using $k$-means with the Euclidean distance metric. Finding that the Euclidean metric outperformed the IoU metric was initially unexpected due to the tendency of Euclidean distance to minimise error for larger bounding boxes at the expense of smaller bounding boxes \cite{yolov2}. Observing the distribution of centroids in Figure \ref{fig:EDA_box} for both $k$-means attempts however, reveals a compliance with a peripheral test conducted in the original Faster R-CNN paper \cite{fasterrcnn}. This highlighted the importance of variety in anchor box scales as opposed to aspect ratios. Though there is greater aspect ratio variation when using the IoU metric, Euclidean distance provides centroids with a greater spread of scales over the groundtruth bounding boxes of the dataset. Though ResNet 101 with manually selected anchors marginally had the highest AP for the \textit{tree\_apple} class at 0.6967, it did also have the widest discrepancy out of the three methods which considered the groundtruth boxes. Its \textit{ground\_apple} AP was only 0.3451, leading to a calibrated mAP of 0.6685. This was not as high as for $k$-means (Euclidean metric) which achieved 0.6707.

\subsection{Box Proposals} 

\begin{figure}[H]
    \centering
    \includegraphics[scale = 0.298]{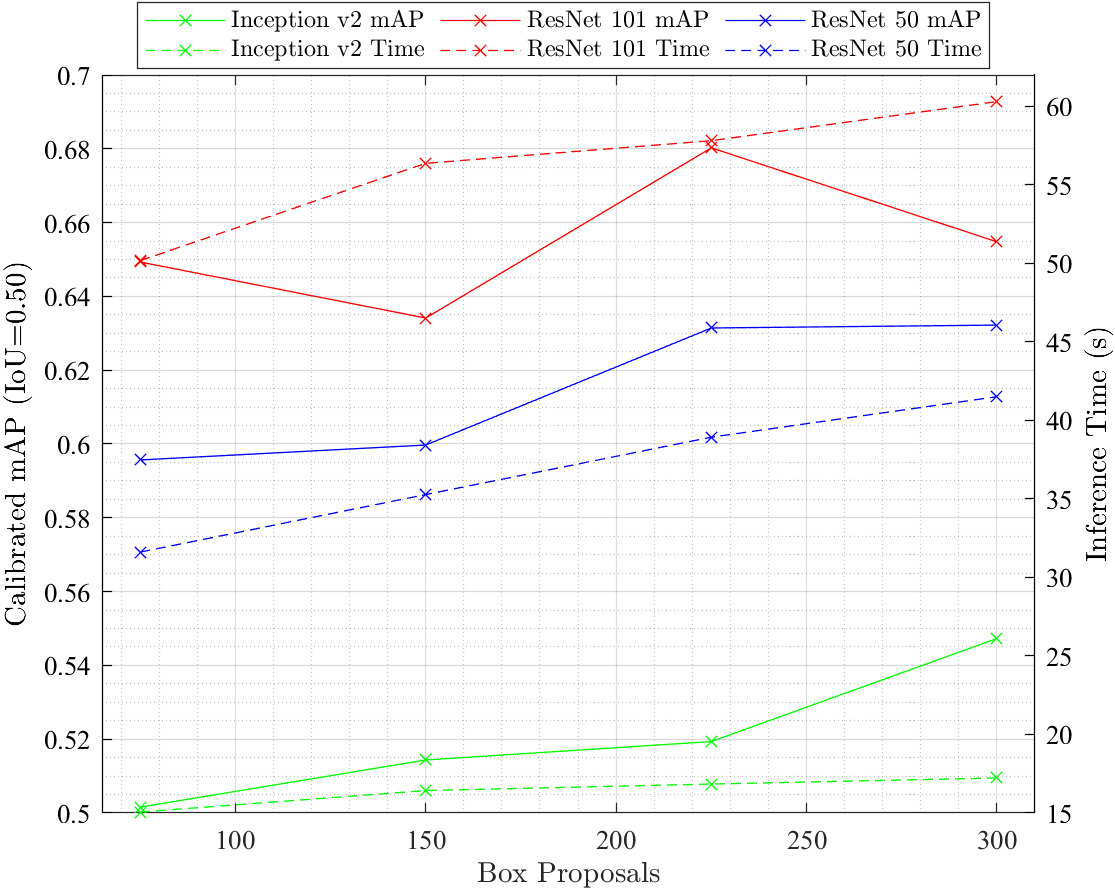}
    \caption{Calibrated mAP and Inference Time (s) As Maximum Number of Box Proposals Varies Between 75 and 300.}
    \label{fig:box_proposals}
\end{figure}

The maximum number of box proposals generated by the RPN was varied from the baseline configuration of 150 proposals. Figure \ref{fig:box_proposals} shows the effect of varying the maximum number of box proposals between 75 and 300 on calibrated mAP and inference time for each of the three feature extractors \cite{smallobjectdetection}.

It can be seen that inference time increases proportionally with the number of box proposals. This increase is intuitive due to the increase in computation, particularly in the costly box-classifier portion of the network. In general, doubling the maximum number of box proposals was found to increase inference time by approximately 5\%, 7\% and 16\% for Inception v2, ResNet 101 and ResNet 50, respectively. More pertinent is the positive correlation between calibrated mAP and maximum number of box proposals for each of the feature extractors. This is shown to increase over the range of 75 to 300 proposals for Inception v2 and ResNet 50. This is in accordance with \cite{vgg16} which also reported 300 proposals to be optimal before accuracy gains plateau whilst computational cost continues to increase.
Notable in this experiment however, was that calibrated mAP for ResNet 101 actually decreased between 225 and 300 box proposals, with drops in AP for both apple classes. Repeated retraining confirmed this finding not to be erroneous. Though this has not been reported in literature surrounding Faster R-CNN, it was notably documented in \cite{fastrcnn} which, as previously stated, operates a selective search algorithm. Their results found that AR, which usually correlates well with mAP, did not correlate when number of box proposals increased. The same was found in this experiment. As box proposals increased from 225 to 300 for ResNet 101, AR (with IoU threshold averaged over 0.50 to 0.95 in 0.05 increments) increased from 0.361 to 0.393, whilst calibrated mAP fell from its highest value, 0.6802, to 0.6548. The consequence of more box proposals for ResNet 101 is therefore seen to be an increase in false positives. Such a pattern was empirically attributed to the density of box proposals saturating mAP in \cite{fastrcnn}.

\subsection{Optimiser}

The training configuration formed the next set of experiments. Three of the most efficient algorithms for gradient descent optimisation were investigated. These are Momentum, a variant of SGD; Adam, an adaptive learning rate algorithm; and RMS Prop, another adaptive method, which divides the learning rate by an exponentially decaying average of squared gradients. The baseline learning rate, $\eta = 0.0003$, was originally instated due to commonly being used with the baseline Momentum optimiser in literature \cite{imagenet}. Similarly, a momentum value, $\gamma=0.9$, was chosen as it features in almost all literature involving Momentum \cite{fasterrcnn, tinyfaces, inceptionv2}. The omission of a fixed learning rate schedule was also elected. This is because it has been found previously that Momentum, when training the Oxford-IIIT Pets set over 200,000 steps on Faster R-CNN \cite{modelzoo}, converges sufficiently. Moreover, adaptive learning rate schedules do not typically require manually decaying learning rates.

\begin{figure}[H]
    \centering
    \includegraphics[scale = 0.34]{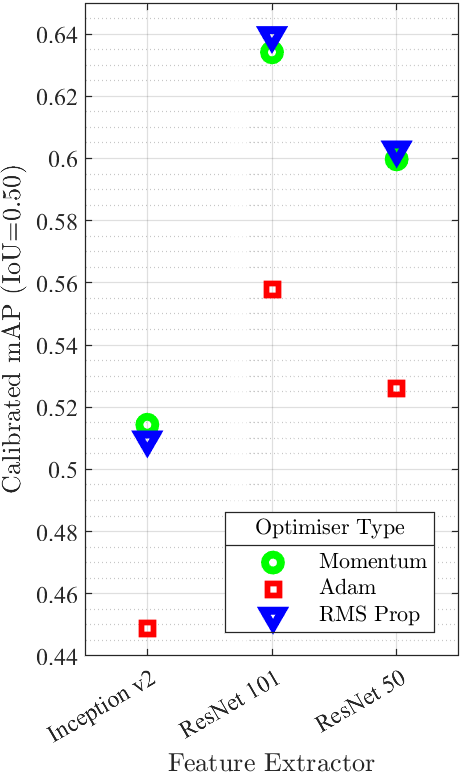}
    \caption{Calibrated mAPs For Different Optimisers: Momentum ($\eta = 0.0003$), Adam ($\eta = 0.001$) and RMS Prop ($\eta = 0.001$).}
    \label{fig:optimiser_types}
\end{figure}

The three feature extractor networks were initially trained using each of the three optimisers, all with $\eta = 0.0003$. This was also deemed the appropriate starting point for the two adaptive learning rate methods due to their robustness in making implicit corrections. A second round of training was conducted on these two methods specifically. This time with an initial learning rate set at, $\eta = 0.001$. This value was recommended in the initial paper for Adam \cite{adam} and has been found empirically to be optimal for RMS Prop too \cite{optimiserreviewpaper}. The experiments found that using $\eta = 0.001$ yielded marginally better calibrated mAP compared to $\eta = 0.0003$ for the two adaptive learning rate methods. As such, Figure \ref{fig:optimiser_types} shows the effect of optimiser choice where Momentum ($\eta = 0.0003$), Adam ($\eta = 0.001$) and RMS Prop ($\eta = 0.001$) are used to train the three feature extractors.

In spite of using the recommended learning rate for Adam, it lead to a reduction in calibrated mAP for each of the three feature extractors, compared with the Momentum baseline. When studying the multitask loss in the RPN, it was found that, whilst bounding box regressor (localisation) loss was similar compared to the equivalent network trained with Momentum, the objectness loss was consistently twice as large for all networks. This fed through to the final classifier loss: localisation losses remained similar and the classification loss disparity between Adam and Momentum remained, though was less extreme. Although Adam has often been noted in literature as being an efficient optimiser \cite{optimiserreviewpaper}, it has also been shown to have poor generalisation capabilities, compared to SGD-based approaches, for certain applications. The results in this work were in agreement with those found in \cite{optimisergeneralistationissues}, which highlighted that for binary classification problems, Adam finds drastically different solutions to SGD-based approaches and has a much higher test error.

The similarity in calibrated mAP performance between Momentum and RMS Prop is mimetic of their multitask losses. These were found to be almost identical for all three feature extractors. It is observed that the RMS Prop optimiser slightly improved mAP for ResNet architectures but lightly reduced it for the Inception v2. Further experimentation on RMS Prop, used as an attempted improvement to Inception v2, is presented in \cite{inceptionv2}. Looking at individual class AP, it was also found empirically that using RMS Prop lead to considerably higher AP for the \textit{ground\_apple} class but a marginal reduction in AP for the \textit{tree\_apple} class. The highest calibrated mAP found (0.6396) was obtained by ResNet 101 which used the RMS Prop optimiser with $\eta = 0.001$.

\subsection{Combining Inception and ResNet Architectures}

Thus far, Inception and ResNet feature extractors have been analysed separately. This experiment evaluates the effect of combining the Inception network with residual connections found in ResNet, as explained in Section \ref{proposedsolutions}. Also justified in that section is the use of Inception ResNet v2 Atrous, a base network which is seldom studied in literature.

When trained with the same baseline hyperparameters as in Table \ref{tab:baseline_params}, the calibrated mAP was found to be 0.6339 with \textit{ground\_apple} and \textit{tree\_apple} APs of 0.3326 and 0.6603, respectively. This is almost identical to the performance of ResNet 101, in spite of what was reported in \cite{modelzoo}. This saw an increase in mAP (over IoU thresholds in the range 0.50 to 0.95) by 0.05 between the two base networks trained on COCO. Less surprising was the 106.87s inference time, which is twice that of ResNet 101.

Potential explanations as to why Inception ResNet v2 Atrous did not yield the desired mAP increase could be attributed to its receptive field size which is three times larger than that of ResNet 101 \cite{inceptionresnet}. This has been shown to deliver state of the art accuracy in datasets with large objects, though, for this application, which consists of only small ($<32^2$ pixels) and medium ($32^2<96^2$ pixels) objects, the receptive field is too large, relatively. Such a phenomenon has been noted in \cite{receptivefield} which saw a reduction in AP for small objects in networks with larger receptive fields.

\subsection{Effect of More Training Data}

\begin{figure}[H]
    \centering
    \includegraphics[scale = 0.315]{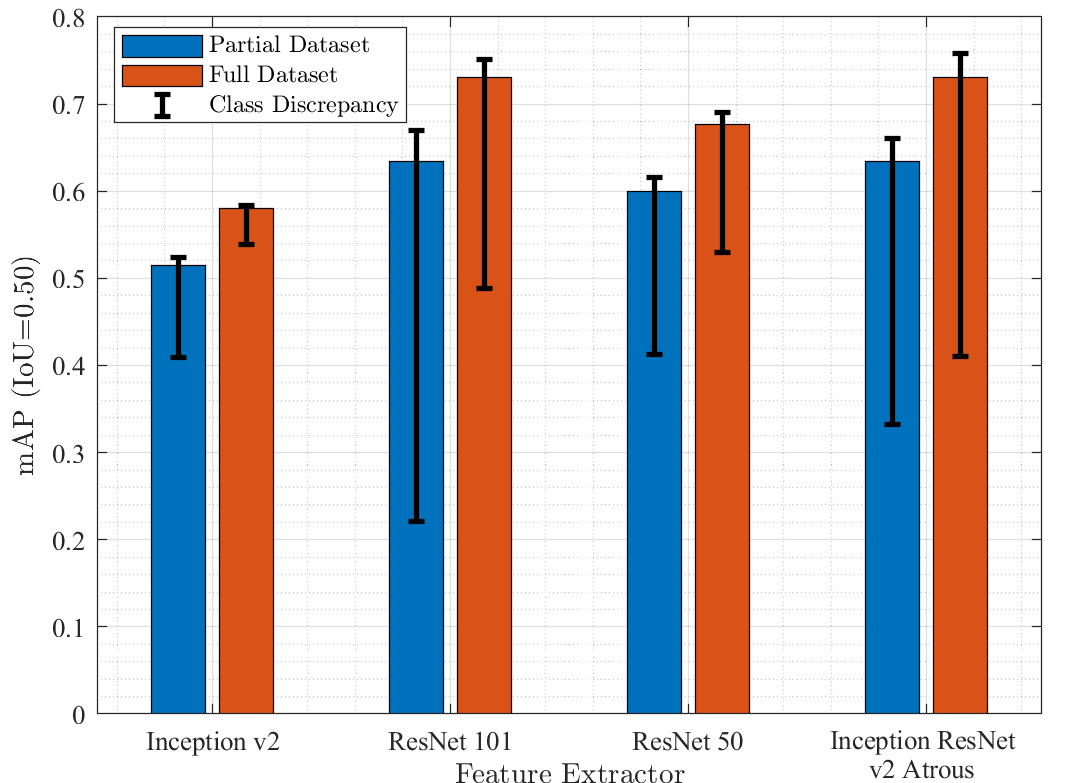}
    \caption{mAPs For Each Feature Extractor Trained on the Partial and Full Datasets. Bars Represent Calibrated mAP. Error Bars Denote the Class Discrepancy: AP of the \textit{tree\_apple} Class (Upper Bound), AP of the \textit{ground\_apple} Class (Lower Bound).}
    \label{fig:x4data_bars}
\end{figure}

A final experiment was conducted to quantify the increase in mAP due to increasing the size of the training dataset. This is of particular interest to those, such as Outfield, who have intentions to extend the object detector to learn different types of fruit. Therefore, knowledge of the quantity of training data required is beneficial.
As stated in Section \ref{experimental_procedure}, training up until this point has been performed on the partial dataset consisting of 107 images and 2,709 marked-up apples, due to time constraints. The effect of increasing the dataset to 435 images and 13,439 marked-up apples is now explored. The results for the feature extractors, including Inception ResNet v2 Atrous, are shown in Figure \ref{fig:x4data_bars}.

As expected, calibrated mAP gains were seen for each case. On average, mAPs experienced a percentage increase of 13\% upon approximately quadrupling the dataset. Also noticeable was the shrinking of the class discrepancy caused by significant gains in AP for \textit{ground\_apple} class. This confirms previous suggestions that the \textit{ground\_apple} class AP was sporadic due to number of marked-up examples in the training dataset.

\subsection{Optimal Model Configuration}

\begin{figure*}[!b]
    \centering
    \includegraphics[scale = 0.2018]{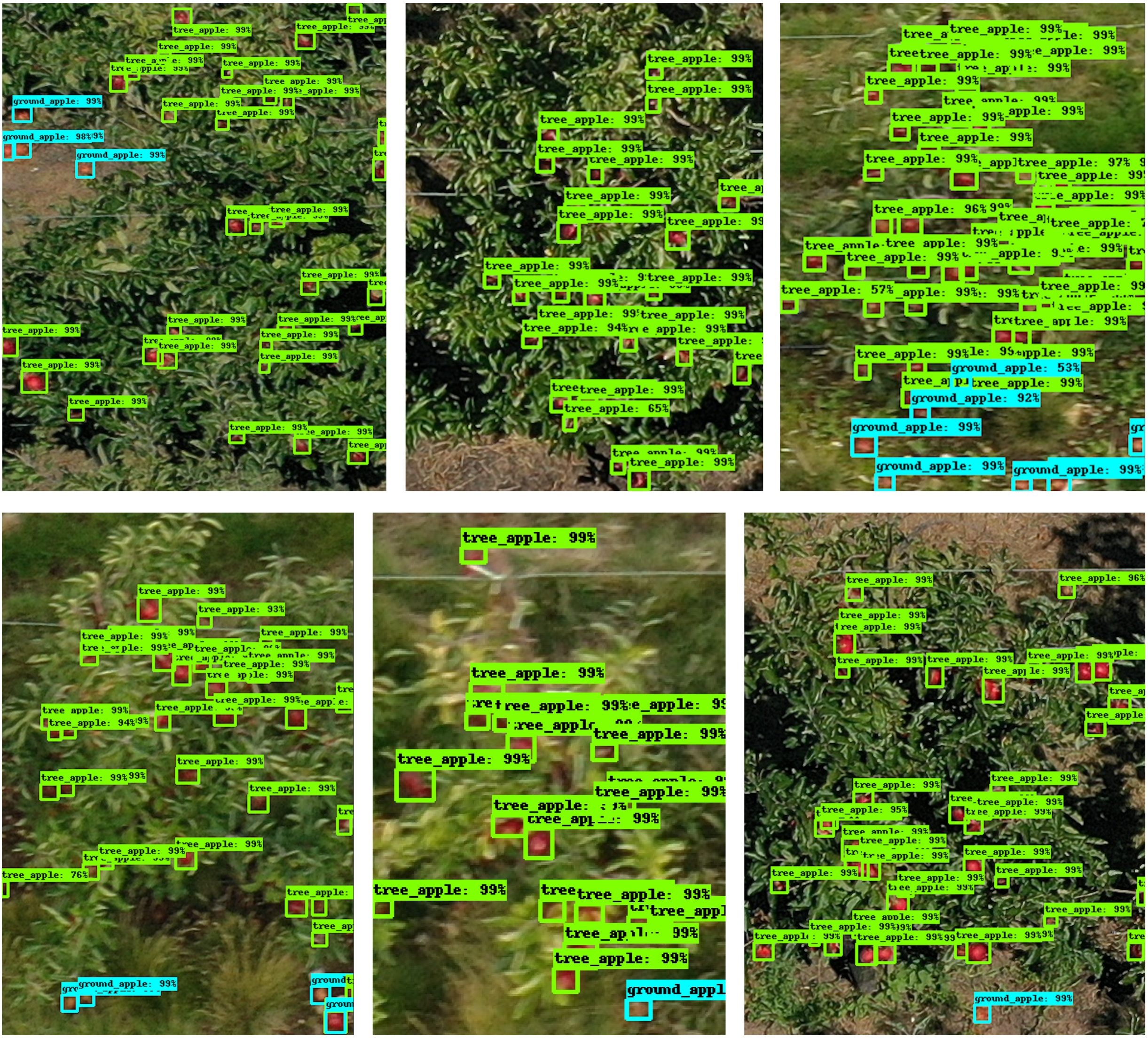}
    \caption{Inference Run on A Sample of Six Images From the Test Dataset. The Two Classes Are Shown In Green and Blue, With Respective Confidences Given For Each Apple.}
    \label{fig:test_images}
\end{figure*}

Based on the results discussed in this section, the optimal model configuration is as follows. A ResNet 101 base feature extractor is selected after being found consistently to attain higher calibrated mAP than its 50-layer variant and the Inception v2 network while hyperparameters were tuned. It was shown to match the performance of the Inception ResNet v2 Atrous network, though its inference time was found to be half, making it preferable. The set of $k=5$ anchor boxes, determined from $k$-means clustering (with the Euclidean distance metric) using bounding box EDA, is selected. Also in the RPN, a maximum of 225 box proposals is permitted. Finally, for training, an RMS Prop optimiser, with $\eta=0.001$ is employed. Remaining hyperparameters are unchanged as in Table \ref{tab:baseline_params}. This optimal model was trained to 200,000 steps on the full 435-image dataset. The performance is summarised in Table IV. 


\begin{figure}[H]
    \centering
    \includegraphics[scale = 0.101]{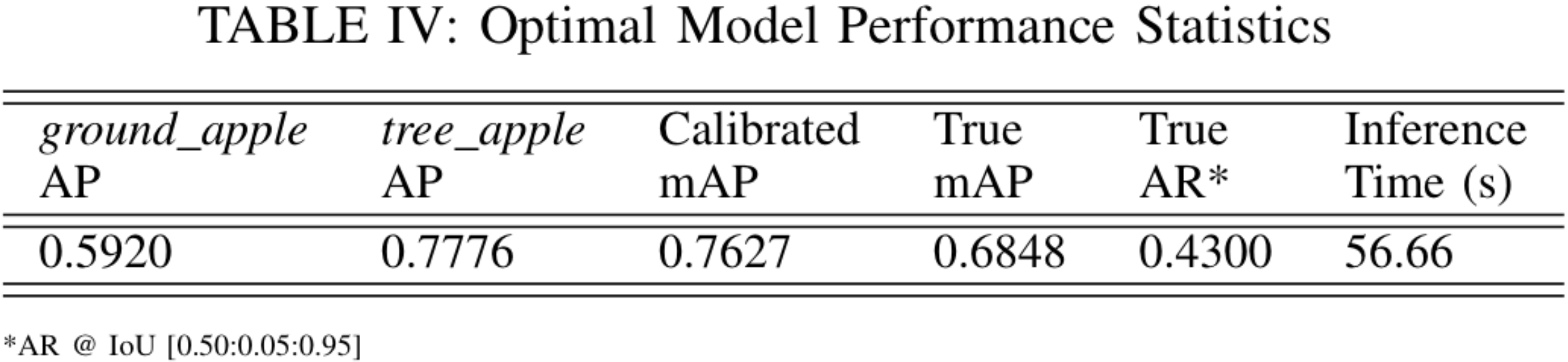}
    \label{fig:table_shrunk}
\end{figure}

This model achieves the best calibrated mAP, individual class APs, true mAP and AR compared to all other models tested. Whilst having a similar inference time as the baseline ResNet 101, its calibrated mAP is 0.1286 higher. Comparing this to \cite{fasterrcnn}, the best Faster R-CNN incarnation from  was found to have a mAP of 0.7040 on the PASCAL VOC 2012 set. This comparison should be treated with caution due to the vastly different datasets however. PASCAL VOC contains 20 classes and large unobstructed objects. Whereas, this dataset contains 2, difficult to distinguish classes, consisting of small, often covered or clustered objects, which have been marked-up to the best possible standard in spite of the unavoidable subjectivity of the dataset, as predicted in Section \ref{experimental_procedure}.

A sample of six separate test photos during inference are shown in Figure \ref{fig:test_images}. These highlight the success of this model in differentiating between the two apple classes. This instils confidence that such a network could easily be extended to detect a wide range of different fruits and potentially even different fruit varieties as soon as training data become available.

\section{Perspective Projection Image Preprocessor System}\label{sec:deployment}

Due to the desire to have tree-level crops as the input to the network, as described in Section \ref{experimental_procedure}, an image prepossessing step must be devised. One method of extracting such crops from the raw, drone images is to pass them through a preliminary NN for detecting or segmenting trees. Such a network has been trained by \cite{hawthorne}. This would however need to be extended to crop out the detected trees before passing them into the proposed networks in this paper. Moreover, there is the issue of double-counting trees, as successive drone images have a specified overlap to improve key-point matching for creating orthomosaics and point clouds. 

A more robust solution, given Outfield's orchard surveying procedure, would be to incorporate the Real-Time Kinematic (RTK) positioning data. This can be collected easily for the bases of trees at the end of each row of the orchard using an RTK receiver. In QGIS, these coordinates can be extrapolated for all the trees using a known tree spacing (provided by the grower), as well as the obtained Digital Terrain Model (DTM). In addition, the Digital Surface Model (DSM) is used to find the height of each tree at these base coordinates. The DTM and DSM are processed using Pix4D following the drone survey. The result is a set of 3D coordinates (latitude, longitude and altitude) for the top and bottom of every tree in the orchard, given in the world reference frame, using the Pseudo-Mercator Projection WGS-84. 

\begin{figure}[H]
    \centering
    \includegraphics[scale = 0.146]{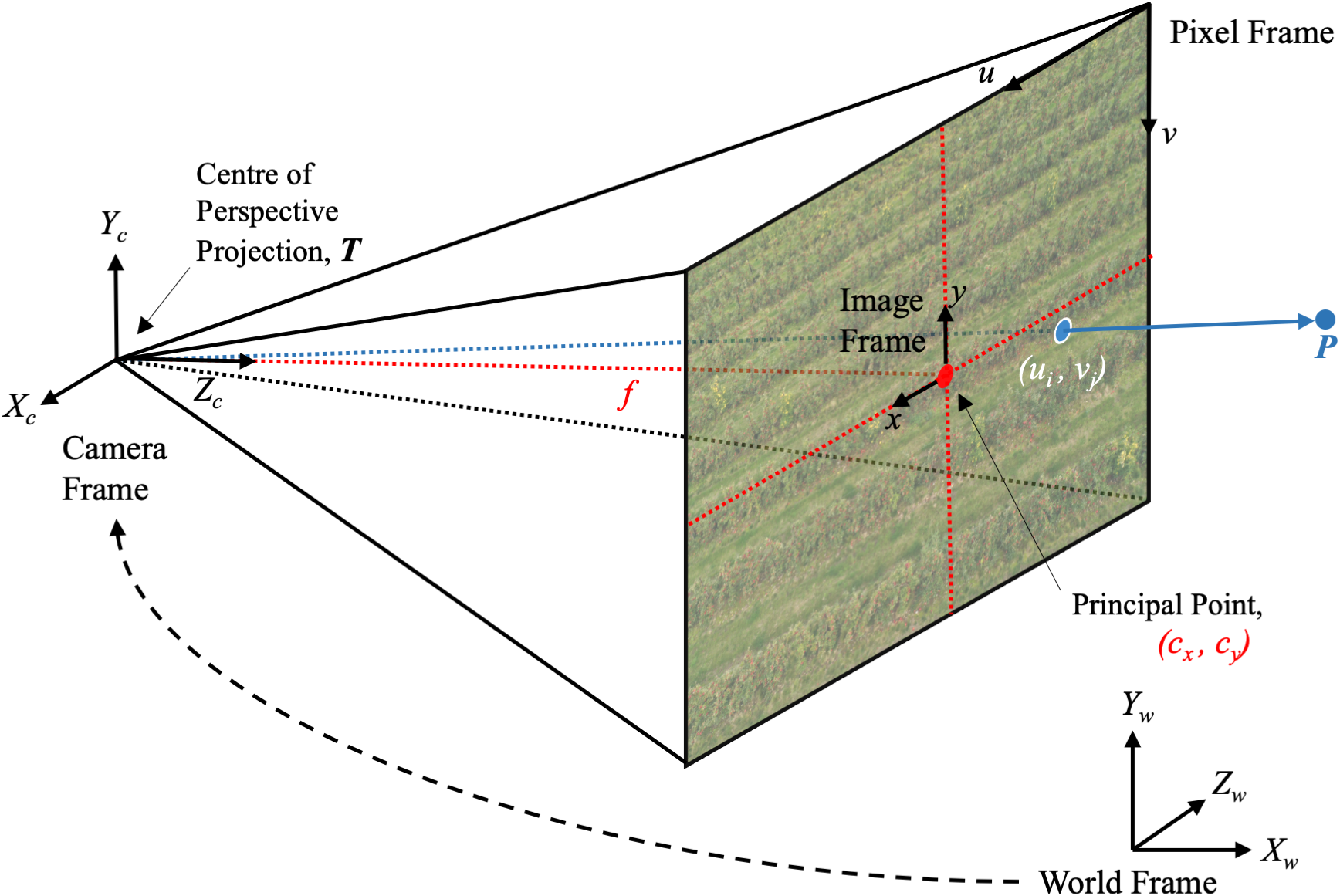}
    \caption{Perspective Projection 3D Geometry For A Camera Without Distortion.}
    \label{fig:fwd_proj}
\end{figure}

To convert these into 2D pixel coordinates, the perspective projection model must be applied. To do this, two types of parameters need to be obtained, namely, the extrinsic and intrinsic camera parameters. The former define the transformation between the camera reference frame and the world frame, whilst the latter relates the pixel coordinates of an image point to the corresponding coordinates in the camera reference frame. A depiction of the perspective projection model using a standard drone image is shown in Figure \ref{fig:fwd_proj}.

Finding the extrinsic camera parameters involves first determining the 3D rotation matrix that brings the axes of the camera frame and the world frame into alignment. Using the angles $\omega$, $\phi$ and $\kappa$, this is shown in,

\begin{equation}
\pmb{R} = \pmb{R}_{x}(\omega) \pmb{R}_{y}(\phi) \pmb{R}_{z}(\kappa)
\label{rotation1}
\end{equation}

which is the multiplication of the 3D rotation matrices, using the right hand rule, as in:

\begin{equation}
\pmb{R}_{x}(\omega) = 
\begin{pmatrix} 1 & 0 & 0 \\ 0 & \cos(\omega) & -\sin(\omega) \\ 0 & \sin(\omega) & \cos(\omega) \end{pmatrix},
\label{rotmat1}
\end{equation}

\begin{equation}
\pmb{R}_{y}(\phi) = 
\begin{pmatrix} \cos(\phi) & 0 & \sin(\phi) \\ 0 & 1 & 0 \\ -\sin(\phi) & 0 & \cos(\phi) \end{pmatrix},
\label{rotmat2}
\end{equation}

\begin{equation}
\pmb{R}_{z}(\kappa) = 
\begin{pmatrix} \cos(\kappa) & -\sin(\kappa) & 0 \\ \sin(\kappa) & \cos(\kappa) & 0 \\ 0 & 0 & 1 \end{pmatrix}.
\label{rotmat3}
\end{equation}

Secondly, the translation between the relative positions of the origins of the world frame and the camera frame must be ascertained. Combining both transformations results in,

\begin{equation}
\pmb{P}_{c} = \pmb{R}^T(\pmb{P}_{w} - \pmb{T})
\label{rotmatcondensed}
\end{equation}

where,

\begin{equation}
\pmb{P}_{c} =
\begin{pmatrix} X_{c} \\ Y_{c} \\ Z_{c} \end{pmatrix} 
, \;
\pmb{P}_{w} = 
\begin{pmatrix} X_{w} \\ Y_{w}\\ Z_{w} \end{pmatrix} 
, \;
\pmb{T} = 
\begin{pmatrix} T_{x} \\ T_{y}\\ T_{z} \end{pmatrix}.
\label{rotmatmeaning}
\end{equation}

$\pmb{P}_{c}$ and $\pmb{P}_{w}$ are the coordinates of a 3D point, $\pmb{P}$, in the camera frame and the world frame, respectively. Whereas, $\pmb{T}$ is the coordinates of the centre of perspective projection in the world frame.

The intrinsic camera parameters are defined by the optical, geometric, and digital characteristics of a camera. These are then found, by Thales' theorem, using the focal length, $f$, the principal point, $(c_{x}, c_{y})$, and perspective projection using the camera frame in,

\begin{equation}
\begin{pmatrix} u_{i} \\ v_{j} \end{pmatrix} 
= 
\begin{pmatrix} \frac{fX_{c}}{Z_{c}} \\ - \frac{fY_{c}}{Z_{c}} \end{pmatrix} 
+
\begin{pmatrix} c_{x} \\ c_{y} \end{pmatrix}
\label{pixelcoords}
\end{equation}

where $u_{i}$ and $v_{j}$ are the pixel coordinates of the 3D point projection using a non-distorted camera model \cite{nondistortedcameramodel}. The code \textit{3D\_2D\_bounding\_crop.py} was written to perform these transformations using the NumPy and OpenCV libraries. It first decodes the project files produced in Pix4D, such as \textit{pmatrix.txt} and \textit{offset.xyz}. Then, for every image in the \textit{undistorted\_images} directory, the code iterates through all the tree coordinates, converting them into the pixel frame, to determine which trees are visible. From these, if a tree has not yet been recorded in a previous image, it is cropped and saved with its uniquely assigned identifier, before being fed into the detection NN. The identifier is a concatenated row and column address based on the real-world position of the tree in the orchard. 

\begin{figure*}[h]
    \centering
    \includegraphics[scale = 0.23]{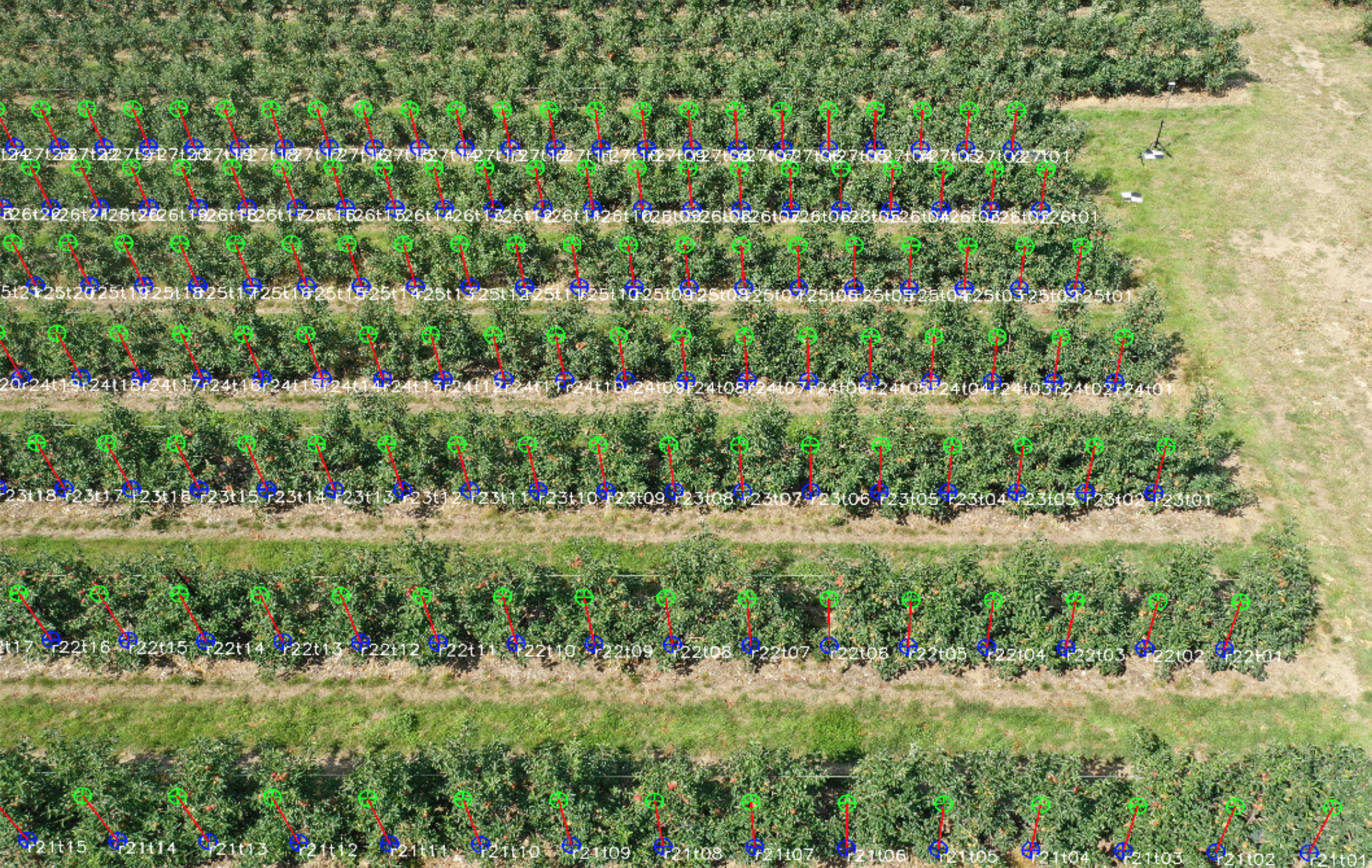}
    \caption{For Illustration: A Raw Drone Image, Where Each Tree Has Been Tagged and Identified Using the Perspective Projection Model. This Is Prior to Being Cropped to the Tree-Level.}
    \label{fig:preprocessorELS}
\end{figure*}

It is this algorithm that makes this system superior to simply cropping all trees from a precursory NN. This is not only due to the aforementioned double-counting issue, but also because of the extra geospatial information provided. Specifically, being able to determine how many apples are growing on each tree, and, crucially, where those trees are located in the orchard. This type of georeferencing is beneficial because it provides data regarding which trees or areas of the orchard are not meeting yield expectations. Physical investigations in the orchard can then be conducted in problem areas, focusing on deficiencies in soil water content, minerals or sunlight, for example. For illustration, a raw drone image, where the identifiers have been automatically generated and superimposed, is shown in Figure \ref{fig:preprocessorELS}.

\section{Conclusions}\label{sec_concl}

This work proposed a system, based on the principles of object detection, to detect and count apples from oblique, overhead drone imagery of giant commercial orchards. As a precursor, a novel technique for extracting individual trees from the raw drone imagery was devised to avoid wasted computation in the NN caused by feeding in such imagery. This incorporated RTK data and the DTM to locate global coordinates of every tree base in the orchard in WGS-84. Then, using forward projection, with the internal and external camera parameters, the set 2D pixel coordinates corresponding to each tree in the imagery could be found. Tree-level crops were then taken, based on the DSM and a known tree spacing, such that every tree was extracted once only. Unique, geospatial identifiers localise yield information to the tree-level, allowing farmers to take further action based on underperforming trees or areas of the orchard.

The tree-level crops obtained from this initial stage formed the set of input imagery to be fed into the detection NN. Optimisation of this network to detect individual apples, and differentiate between apples on trees and on the ground, was then conducted. 
From previous works, Faster R-CNN was established as the most suitable object detection framework due to its region proposal approach which prioritises accuracy over training and inference time, unlike regression-based approaches. Three base feature extractors formed the basis on which network hyperparameters were tuned. Parameters whose adjustment resulted in positive increases in calibrated mAP were reported. Anchor box design was key due to the atypical scale of apple objects. A $k$-means clustering approach based on Euclidean distance and EDA of groundtruth bounding boxes yielded the biggest increase in calibrated mAP compared to the baseline anchors. Also explored were the number of box proposals in the RPN, where mAP was found to peak or stagnate at 225 proposals for the ResNet bases. Detrimental performance above this was attributed to an increase in false positive detections, which reduced mAP whilst AR continued to rise. Experiments on the training configuration explored the effect of optimiser types and learning rate combinations. Adam optimiser was found to be significantly less successful than what was reported in literature due to poor generalisation capabilities. RMS Prop, with $\eta=0.001$, was found to be a marginal improvement on Momentum. Final experiments showed that combining Inception and ResNet blocks did not give rise to the expected increase in calibrated mAP compared to ResNet 101. This was potentially due the difference in size between network receptive field and apple objects. Inference time also doubled, making it unfavourable compared to ResNet 101. Lastly, quadrupling the dataset size was quantified by an approximate 13\% increase in calibrated mAP.

The amalgamation of the optimal hyperparameters lead to a model configuration with a calibrated mAP of 0.7627 and a true mAP of 0.6848, both of which sit among state of the art figures quoted in literature, though for drastically different detection tasks. The ability of the network to distinguish between ground apples (AP = 0.5920) and tree apples (AP = 0.7776) was particularly impressive. This leads to confidence that this network can successfully be extended to different types and varieties of fruit when training data become available.

\section*{Acknowledgement}

The authors would like to thank Outfield Technologies (Oli Hilbourne and Jim McDougall) for the data and motivation behind the project.

\bibliography{finyearprjbib}

\end{document}